\documentclass[journal,twoside,web]{ieeecolor}
\usepackage{generic}
\usepackage{cite}
\usepackage{amsmath,amssymb,amsfonts}
\usepackage{algorithmic}
\usepackage{graphicx}
\usepackage{textcomp}
\usepackage{multirow}
\usepackage{lscape}
\usepackage{booktabs}
\makeatletter
\def\fnum@figure{\textcolor{subsectioncolor}{\sf Fig.~\thefigure}}
\def\fnum@table{\textcolor{subsectioncolor}{\sf TABLE~\thetable}}
\makeatother
\newcommand{\norm}[1]{\left\lVert#1\right\rVert}
\def\BibTeX{{\rm B\kern-.05em{\sc i\kern-.025em b}\kern-.08em
    T\kern-.1667em\lower.7ex\hbox{E}\kern-.125emX}}
\begin{document}
\title{C$^2$SP-Net: Joint Compression and Classification Network for Epilepsy Seizure Prediction}
\author{Di Wu$^*$, Yi Shi$^*$, Ziyu Wang, Jie Yang, and Mohamad Sawan, \IEEEmembership{Fellow, IEEE}
\thanks{Manuscript submitted xxx.}
\thanks{
Di Wu is with Zhejiang University, Hangzhou 310007, China and also with the Center of Excellence in Biomedical Research on Advanced Integrated-on-chips Neurotechnologies (CenBRAIN Neurotech), School of Engineering, Westlake University, Hangzhou 310024, China.}
\thanks{
Yi Shi, Ziyu Wang, Jie Yang, and Mohamad Sawan are with the Center of Excellence in Biomedical Research on Advanced Integrated-on-chips Neurotechnologies (CenBRAIN Neurotech), School of Engineering, Westlake University, Hangzhou 310024, China, and also with the Institute of Advanced Technology, Westlake Institute for Advanced Study, Hangzhou 310024, China (e-mail:  yangjie@westlake.edu.cn, sawan@westlake.edu.cn).}
\thanks{
This work was supported by Zhejiang Key R\&D Program project No.
2021C03002, and Zhejiang Leading Innovative and Entrepreneur Team Introduction Program
No. 2020R01005.}
\thanks{* Indicates equal contribution.}
}

\maketitle
\begin{abstract}
Recent development in brain-machine interface technology has made seizure prediction possible.  However, the communication of large volume of electrophysiological signals between sensors and processing apparatus and related computation become two major bottlenecks for seizure prediction systems due to the constrained bandwidth and limited computation resource, especially for wearable and implantable medical devices. Although compressive sensing (CS) can be adopted to compress the signals to reduce communication bandwidth requirement, it needs a complex reconstruction procedure before the signal can be used for seizure prediction. In this paper, we propose C$^2$SP-Net, to jointly solve compression, prediction, and reconstruction with a single neural network. A plug-and-play in-sensor compression matrix is constructed to reduce transmission bandwidth requirement. The compressed signal can be used for seizure prediction without additional reconstruction steps. Reconstruction of the original signal can also be carried out in high fidelity. Prediction accuracy, sensitivity, false prediction rate, and reconstruction quality of the proposed framework are evaluated under various compression ratios. The experimental results illustrate that our model outperforms the competitive state-of-the-art baselines by a large margin in prediction accuracy. In particular, our proposed method produces an average loss of 0.6\% in prediction accuracy with a compression ratio ranging from 1/2 to 1/16.
\end{abstract}

\section{Introduction}
\label{sec:introduction}
\IEEEPARstart{E}{pilepsy} is a neurological disease that causes recurrent seizures in the brain, influencing the lives of over 50 million people\cite{Fisher2014}\cite{zhang2016}. Patients experience unconsciousness, movement disorders, and other control loss of body parts when a seizure occurs. Nowadays, many wearable and implantable circuits and systems\cite{Yang2020} are developed to detect\cite{6634266} or predict\cite{BouAssi2018} the occurrence of seizures to provide time for emergent preparation under risky scenarios without medical attendance, such as driving and operating heavy machines.

\begin{figure}[t]
\centerline{\includegraphics[width=\columnwidth]{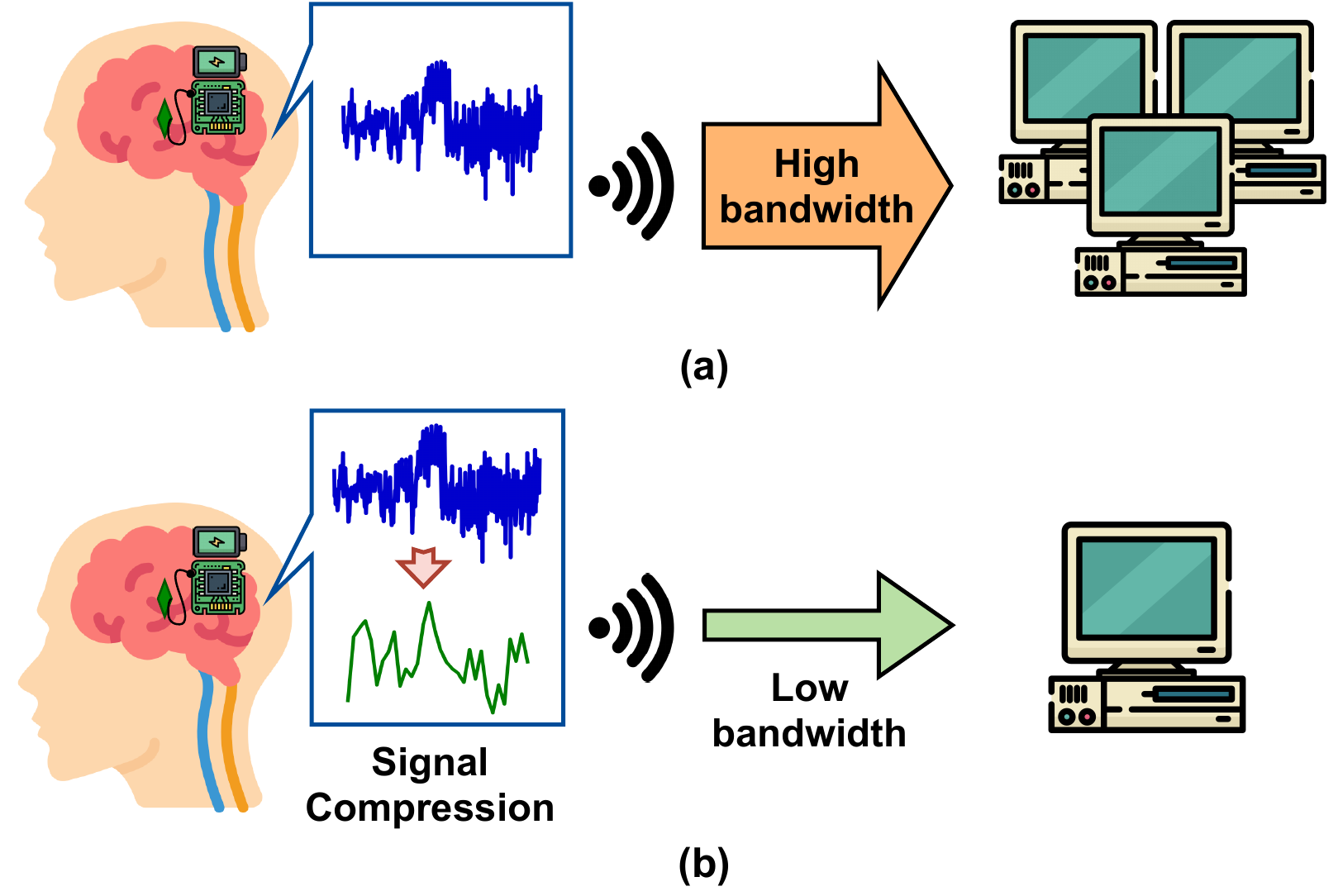}}
\caption{Comparison of the signal acquisition, transmission and analysis process between conventional system and system with in-sensor signal compression. }
\label{fig_intro}
\end{figure}

Fig. \ref{fig_intro}-(a) shows the system built-up of commonly seen seizure prediction systems. It usually consists of electrodes to collect electrophysiological signals, which reflect the brain's neuron activities. The signals are further transmitted to a processing apparatus such as a micro controller unit (MCU) and dedicated signal processors through weak and short distance RF communication \cite{7433928}.

Due to the recurrent nature of epilepsy, these systems are required to be wearable or implantable; hence the communication bandwidth and computation ability are both limited due to the miniature and power constraints. In-sensor signal compression is therefore essential to ensure the usability of wearable/implantable real-time prediction systems \cite{Yang2020}. As illustrated in Fig. \ref{fig_intro}-(a), the conventional electrophysiological signals analyzing process requires devices to have high bandwidth in data transmission and massive computation in subsequent analysis.

In-sensor signal compression could vastly reduce the transmission bandwidth and computational cost of downstream tasks. Conventional compression methods such as average downsampling \cite{Combaz2009} and uniformly downsampling \cite{Soleymani2013} are largely limited by the Nyquist sampling rate, and the compressed data could not be reconstructed afterward. Compressive sensing (CS) has been proposed to compress signals at a sub-Nyquist sampling rate in recent years. The compression process could be denoted as a matrix multiplication between the signal and a sensing matrix, which is computationally efficient and hardware friendly\cite{Mamaghanian2011}. In addition, CS enables sampling and compression simultaneously, which significantly alleviates on-chip transmission bandwidth and storage space. Fig. \ref{fig_intro}-(b) illustrates how CS methodology is involved in seizure detection systems \cite{7515159}.


However, existing methods \cite{6786004} take compression and reconstruction as two separate processes. We argue that signals compressed (randomized CS measurements) with current CS approaches are feasible for reconstruction under sparsity assumption but might not be suitable for downstream tasks such as seizure prediction. Moreover, typical CS algorithms which widely rely on convex optimization \cite{Morabito2013}, greedy algorithms \cite{Lee2019}, and Bayesian learning \cite{Zhang2013} for reconstruction are computationally expensive, making them impractical under real-time scenarios.


\begin{figure*}[t]
	\centering
	\includegraphics[scale=0.5]{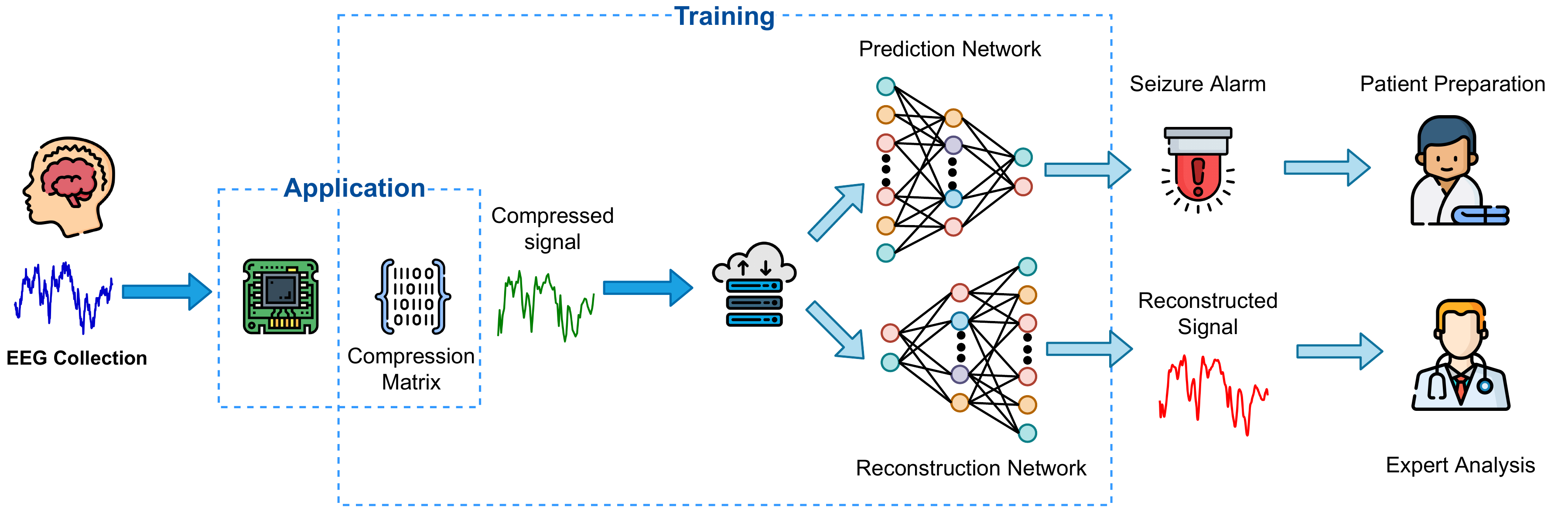}
	\caption{
	A conceptual illustration of our proposed framework. During the training phase, we optimize the compression matrix alongside the reconstruction and prediction network. After training, the obtained compression matrix could then be deployed onto wearable devices for EEG signal compression to reduce transmission cost during actual application.
	}
	\label{fig:system}
\end{figure*}
To mitigate the previously discussed limitations and provide a feasible solution for a reliable and efficient seizure prediction system, we propose a novel end-to-end deep learning framework that jointly solves the signal compression, reconstruction, and seizure prediction tasks. Our proposed framework ensures that the sensing matrix learned is optimized both for reconstruction and prediction purposes. Compared with random sensing matrices, our learned matrix captures informative features based on various downstream tasks. The learned sensing matrix could be deployed onto electroencephalography (EEG) sensors as a general plug-and-play solution for low-cost data compression and transmission. Moreover, the compressed signal can be used directly for seizure prediction or EEG reconstruction. Unlike traditional CS algorithms, Our proposed compression and reconstruction mechanism is parametric and could be directly applied to previously unseen data as opposed to solving an optimization task on new data. To the best of our knowledge, we are the first to explore the feasibility and reliability of CS-involved seizure prediction using EEG. Moreover, besides the application of seizure prediction using scalp EEG (sEEG), our proposed framework can be applied to other wearable and implantable electrophysiological signal application scenarios, such as electrocorticography (ECoG) and local field potential (LFP) to reduce communication bandwidth.

The main contributions of this paper are : 

\begin{itemize}
\item A solution for an efficient and reliable EEG seizure prediction system. We propose a deep learning framework that jointly solves compression, reconstruction, and seizure prediction tasks as a single optimization task.
\item A reconstruction module composed of linear pooling and convolution operation with an adaptive structure under different compression ratios. 
\item A CNN-based seizure prediction network that utilizes ResNet architecture to capture EEG features at various granularity levels.
\item Extensive experiments on popular open-source datasets show that our proposed framework yields new state-of-the-art prediction and reconstruction performance and is stable under different compression ratios.
\end{itemize}

The remainder of this paper is organized as follows. We introduce in section \ref{sec:related_work} previous works in both seizure prediction and compressive sensing areas. In section \ref{sec:method}, we provide the detailed description of this proposed framework. The performance of this proposed framework is evaluated in Section \ref{sec:results}. Finally, the paper is concluded in Section \ref{sec:conclusion}.
\section{Related Work}
\label{sec:related_work}

\subsection{Compressive sensing}
CS was initially brought up for the acquisition of low-rate images, and was gradually developed into various fields including video processing \cite{Mun2011}, face recognition \cite{Nagesh2009} \cite{QIAO2010331}, magnetic resonance imaging (MRI) acquisition \cite{Sandino2020}, etc. CS performs signal compression by sampling a few measurements, i.e. $z = Ax$, where $x \in \mathbb{R}^{N}$ is the original signal, $z \in \mathbb{R}^{M}$ is the sampled measurements (compressed signal) and $A \in \mathbb{R}^{N \times M}$ represents the sampling/sensing matrix with $M < N$.

Most studies involving CS use random matrices as sensing matrices. Zeng \textit{et al.} \cite{ZENG2016497} select the Bernoulli matrix for compressive sensing. They regard EEG compressibility as a kind of feature in terms of seizures detection. They use four classifiers, including decision tree, K-nearest neighbor (K=5), discriminant analysis, and support vector machine (SVM) to classify the features. Through this method, the highest prediction accuracy achieves 76.7\%. Abdulghani \textit{et al.} \cite{Abdulghani2011} use Gaussian matrix to compress the EEG data in their work, and they investigate the performance of different implementations of the CS theory involved in EEG signals. These random matrices fail to embed data-specific or downstream tasks related features into the compressed signals.

The signal reconstruction process from compressed measurements is to solve an optimization task which can be formulated as follows:

\begin{equation}
\underset{x}{\mathrm{min}} \  \mathfrak{R}(x), \ s.t.\ z = Ax,
\end{equation}
where $\mathfrak{R}(\cdot)$ is a regularization term.

To solve the under-determined optimization function, a variety of algorithms have been developed to reconstruct the signals. Among them, Greedy algorithm\cite{BLUMENSATH2009265} is popular due to its low computational complexity. Mallat \textit{et al.} \cite{Mallat1993} came up with a Greedy algorithm called matching pursuit. It decomposes signals into a linear expansion of waveforms that are selected from a redundant dictionary of functions. However, it requires prior knowledge of the sparsity of the underlying signal. Bayesian learning is another algorithm that is common in CS studies. The block sparse Bayesian learning (BSBL) algorithm is an example of Bayesian learning. Initially, the BSBL framework was proposed for signals with a block structure. Zhang \textit{et al.} \cite{Zhang2013} adopt BSBL algorithm in their research, introducing the technique to the telemonitoring of EEG. Their experiment shows good reconstruction quality with an average normalized mean square error (NMSE) of 0.116 and an average structural similarity index measure (SSIM) of 0.81. The Bayesian method has a high speed, but it depends on preliminary knowledge and causes massive computation costs.

To overcome the limitation of conventional CS methods, we propose to construct a sensing matrix during the optimization process of reconstruction and seizure prediction tasks in our framework.

\textcolor{subsectioncolor}{\subsection{Seizure prediction}}
Epilepsy influences 1\% of the world's population, of which up to 35\% could not be cured by pharmaceutical or medical treatment \cite{WHO2006}. Inevitably, these people suffer from unexpected seizure onset. Therefore, effort has been made towards accurate alarm before seizure onset to provide better lives for epileptic sufferers \cite{Xu2020} \cite{Zhang2021} \cite{Lawhern2018}. As an important source for monitoring brain activities in the entire process of epileptic seizure, EEG became the primary focus of the seizure prediction study. Early machine learning based approaches utilize support vector machine (SVM) \cite{Parvez2015} \cite{Yang2018EpilepticSP} and multi-layer perceptrons (MLP) \cite{BEHNAM2016} for seizure prediction. However, these methods rely on hand-engineered features that require a lot of prior knowledge \cite{Mirowski2008}. Park \textit{et al.} \cite{Park2011} introduced SVM to seizure prediction. They calculate spectral power in nine bands from the EEG of the Freiburg EEG database using four pre-processing methods, namely raw, bipolar, time-differential, and bipolar/time-differential. SVM with double cross-validation is applied for classification. Their algorithm results in a prediction sensitivity of 98.3\% and a false prediction rate (FPR) of 0.29/h. However, the manual feature extraction takes a long time, and the features extracted lack generalization ability.

With the successful application of deep learning (DL) methods in many fields, convolutional neural network (CNN) and recurrent neural network (RNN) are extensively used in recent studies~\cite{TSIOURIS201824,wang2020seizure,Zhao2021,Truong2018,rasheed2021generative}. Convolution operations could be seen as filters and thus act as a learnable automatic feature extractor. RNNs are adopted to model the relationship between time sequences better. Truong \textit{et al.} \cite{Truong2018} proposed a generalized retrospective and patient-specific seizure prediction method based on CNN. They utilize Short-time Fourier transform (STFT) to enhance time-frequency information and then apply a CNN-based model for feature extraction and binary classification between pre and interictal states. Their approach achieves an average sensitivity of 81.2\%, and an average FPR of 0.16/h on Children's Hospital of Boston-MIT (CHB-MIT) sEEG database \cite{Shoeb2010}. 

Although CNN models are commonly used in many research activities, some authors claim that RNN might be better at isolating temporal characteristics. Tsiouris \textit{et al.} \cite{TSIOURIS201824} first applied a two-layer long short-term memory (LSTM) network to seizure prediction. Different lengths of preictal windows (ranging from 15 minutes to 2 hours) are used in the seizure prediction task. Prior to classification between preictal and interictal classes, the LSTM model extracts time and frequency domain features between EEG channels cross-correlation and graph-theoretic features. Their results yield a 99.28\% prediction sensitivity and an FPR of 0.11/h on the CHB-MIT sEEG database. 

Despite the success of deep learning based methods on seizure prediction in terms of high prediction accuracy, little to no effort has been dedicated to efficient seizure prediction using compressed EEG data.



\section{Methodology}
\begin{figure*}[t]
	\centering
	\includegraphics[scale=0.35]{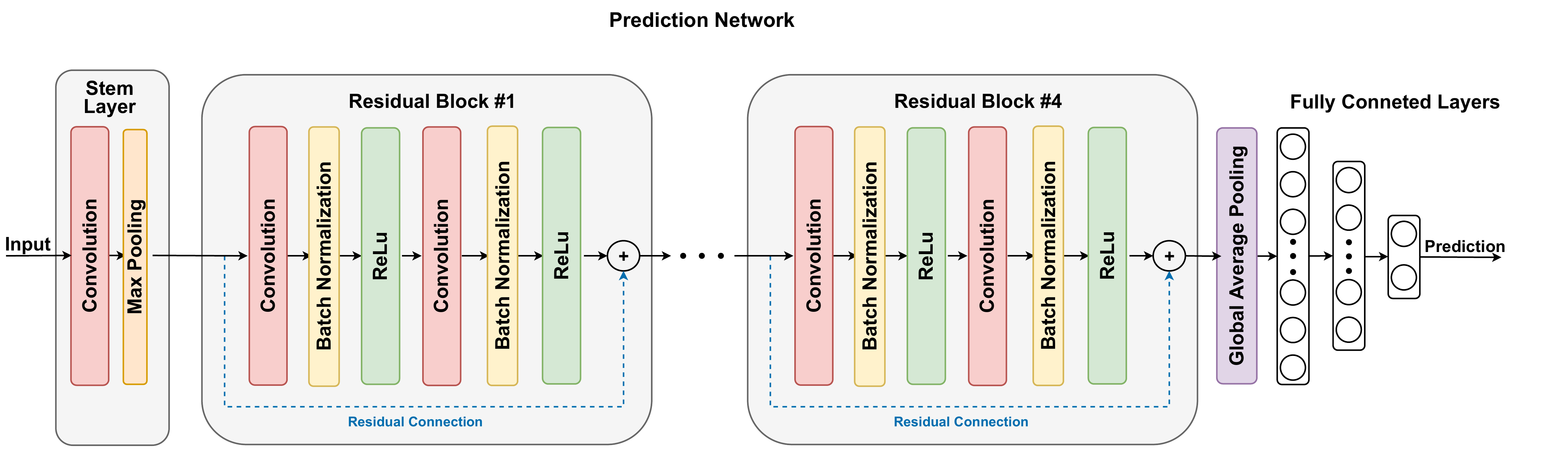}
	\caption{
	Architecture overview of the prediction network. The compressed/original signal is first fed into a stem layer consisting of a convolution layer and a max-pooling layer. Then the output goes through four residual blocks. The dotted blue line indicates the residual connection with dimensional increase. The output of the last residual block is fed into a global average pooling layer and fully connected layers.
	}
	\label{fig:prediction}
\end{figure*}
\label{sec:method}

 In this section, we formalize our proposed C$^2$SP-Net framework, as illustrated in Fig. \ref{fig:system}. The primary purpose of our proposed framework is to compress the original EEG signal for efficient real-time seizure prediction with minimum degradation in prediction performance. Unlike previous CS-based compression methods, which fail to capture signal statistics, our compression strategy embeds prior knowledge of reconstruction and prediction into the compressed signal. Furthermore, our proposed framework is able to reconstruct the original signal for human visualization. The proposed framework contains a compression function $\mathcal{C}(\cdot)$, a reconstruction function $\mathcal{R}(\cdot)$ and a prediction function $\mathcal{P}(\cdot)$. Let $x \in \mathbb{R}^{N \times C}$ be an EEG signal sequence of length $N$ and $y$ be its corresponding one-hot label indicating interictal or preictal state, where $C$ denotes the number of channels. The input signal $x$ is first compressed as follows:
\begin{equation}
z = \mathcal{C}(x),
\label{eq}
\end{equation}
where $z \in \mathbb{R}^{M \times C}$ denotes the compressed signal. Naturally, the signal compression ratio $r$ could be defined as $\frac{M}{N}$ with $M < N$. Then, the reconstructed signal $\hat{x}$ is given by $\mathcal{R}(z)$ while a prediction result $\hat{y}$ is given by $\mathcal{P}(z)$. We design three deep learning based networks to implement these three functions. Next, we describe them in detail. 

\subsection{Compression Network}
Since the purpose of compression is to reduce the computation and transmission cost on wearable/implantable devices, \textcolor{black}{we consider using compression matrix with different bit precision as the compression network. In particular, we choose floating-point compression matrix $W \in \mathbb{R}^{N \times M}$ and binary compression matrix $W \in \{0, 1\}^{N \times M}$.} Then signal from each channel is compressed by multiplying the compression matrix with the original signal, $y^c = Wx^c$, where $y^c$ and $x^c$ denotes the compressed signal and the original signal of channel $c$ respectively. For simplicity, we use the same compression matrix for all signal channels.

\subsection{Prediction Network}
\begin{figure*}[t]
	\centering
	\includegraphics[scale=0.36]{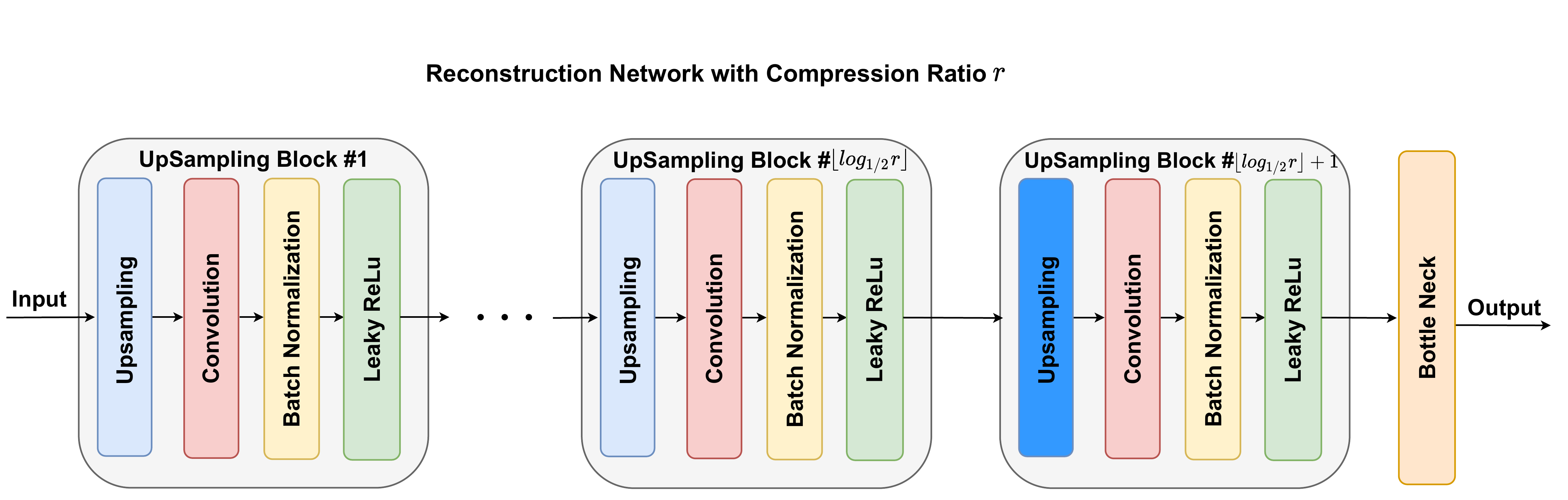}
	\caption{
	Architecture overview of the reconstruction network with compression ratio $r$. It is composed of $\lfloor log_{\frac{1}{2}} r \rfloor + 1$ up-sampling blocks with the $\textit{BottleNeck}$ operation mapping the signal to the original number of channels. The first $\lfloor log_{\frac{1}{2}} r \rfloor$ blocks upsample the signal to twice the length, while the last block upsamples the signal to original signal length.
	}
	\label{fig:recon}
\end{figure*}
\label{subsection: Prediction}
To capture the patterns of different granularity levels from compressed raw EEG signals directly, we designed a CNN based network inspired by the popular ResNet\cite{he2016deep}, as shown in Fig. \ref{fig:prediction}. Specifically, the input signal is first fed into a stem layer which consists of a convolution and a max-pooling operation. Then, the output signal from the stem layer goes through a cascade of basic convolution blocks. Each basic block consists of convolution layers with each followed by a batch normalization layer and an activation layer. In our work, we adopt the rectified linear unit (ReLu) as the activation function. Each basic convolution block contains a residual connection defined as follows:
\begin{equation}
f_{l,k} = \delta(BN(conv_{l,k}(f_{l,k-1}))),
\label{eq2}
\end{equation}
\begin{equation}
f_{l} = f_{l,0} + \textit{BottleNeck}(f_{l,2}),
\label{eq3}
\end{equation}
where $f_{l,k}$ denotes the intermediate feature map of after passing $k^{th}$ convolution layer of block $l$, namely $conv_{l,k}$. $BN$ and $\delta$ stands for batch normalization and activation function receptively. In this paper, we set $k=2$ and $l=4$. The \textit{BottleNeck} operation reduces the number of channels of $f_{l,2}$ to the same number of $f_{l,0}$. The final output $f_{l}$ of block $l$ is then given by addition (residual connection). 

Note that most existing CNN-based approaches treat EEG signal as a greyscale image, meaning the channel dimension and the time dimension of EEG signal are considered the height and width dimension analogous to an image. Thus, 2D convolution is commonly adopted. The convolution operation is considered to be able to extract local spatial information with the assumption that nearby pixels are of high correspondence, which is valid for natural images. However, for the same correspondence to hold for multichannel EEG signals, nearby EEG signals from the channel dimension should be functionally correlated. Unfortunately, due to the different choice of placement of electrodes during EEG signal acquisition, we hypothesize that most existing CNN-based methods fail to capture the discriminative correlation patterns across channels. We seek to capitalize on the interactions among different channels of EEG signal by setting the kernel size of the channel dimension the same as the number of channels of input signal/intermediate feature map, also known as 1D convolution.

The prediction result $\hat{y}$ is then given by:
\begin{equation}
\hat{y} = \sigma(FC_2(\delta(FC_1(GAP(f_{4}))))),
\label{eq4}
\end{equation}
where $GAP$ denotes global average pooling and $FC_i$  denotes the $i^{th}$ fully connected layer. $\delta$ and $\sigma$ represent the LeakyReLu and Softmax activation functions respectively.

We define the prediction loss function as:
\begin{equation}
\mathcal{L}_{pred} = \mathcal{L}(\hat{y}, y),
\label{eq5}
\end{equation}
where $\mathcal{L}$ represents any loss function to evaluate the prediction performance.

\subsection{Reconstruction Network}

We attempt to access the feasibility and reliability of a seizure prediction system using EEG signal under various compression ratios. Besides prediction, we also take reconstruction into consideration for other possible applications which require the original signal such as human expert visualization. Moreover, the reconstruction task could serve as a regularizer for the prediction task to alleviate overfitting.We design a reconstruction network with adaptive architecture according to different compression ratio, as illustrated in Fig. \ref{fig:recon}. We design an up-sampling block defined as follows:

\begin{equation}
\hat{f}_{l} = \delta(\textit{BN}(conv_{l}(\textit{Up}(\hat{f}_{l-1})))),
\label{eq6}
\end{equation}
\begin{table}[ht]
\centering
\caption{Summary of selected patients.}
\label{tab:patient}
\resizebox{\linewidth}{!}{%
\begin{tabular}{@{}cccc@{}}
\toprule
Patient ID & Total seizure & Lead seizure & Sample Count \\ \midrule
chb01      & 7             & 3            & 354           \\
chb05      & 5             & 2            & 353           \\
chb06      & 10            & 6            & 685           \\
chb07      & 3             & 2            & 238           \\
chb08      & 5             & 3            & 355           \\
chb09      & 4             & 3            & 357           \\
chb10      & 7             & 6            & 547           \\
chb14      & 8             & 4            & 456           \\
chb18      & 6             & 3            & 247           \\
chb19      & 3             & 3            & 238           \\
chb22      & 3             & 3            & 301           \\ \bottomrule
\end{tabular}%
}
\end{table}
where $\hat{f}_{l-1}$ and $\hat{f}_{l}$ represents the input feature map and reconstructed intermediate feature map of block $l$ respectively and $\textit{Up}$ denotes up-sampling operation such as linear interpolation. We refer to the operations defined in Section  \ref{subsection: Prediction} for all other unstated notations. In particular, given compression ratio $r$ and original signal length $L$, the reconstruction network is composed of $\lfloor log_{\frac{1}{2}} r \rfloor +1$ pooling blocks. The first $\lfloor log_{\frac{1}{2}} r \rfloor$ blocks each up-samples the signal to twice its input length and the last block restores the signal to the original length. Finally, a $\textit{BottleNeck}$ operation is performed to map the signal to the original number of channels.
We define the prediction loss function as:
\begin{equation}
\mathcal{L}_{recon} = \mathcal{L}(\hat{x}, x),
\label{eq7}
\end{equation}
where $\hat{x}$ represents the reconstructed signal and $\mathcal{L}$ is any differentiable loss function to evaluate the reconstruction quality.

\subsection{Joint-objective Training}
We optimize the compression network, prediction network, and reconstruction network simultaneously using a joint objective function:
\begin{equation}
\mathcal{L}_{joint} = \mathcal{L}_{pred} + \lambda * \mathcal{L}_{recon},
\label{eq8}
\end{equation}
where $\lambda$ is the weighting parameter to balance the two losses. Without loss of generality, we choose cross entropy and mean square error as the loss function for $\mathcal{L}_{pred}$ and $\mathcal{L}_{recon}$, respectively. The compression matrix is obtained by minimizing the joint object and is thus capable of compressing EEG signals in a way for better reconstruction and prediction purposes. 
\begin{figure}[t]
\centerline{\includegraphics[width=\columnwidth]{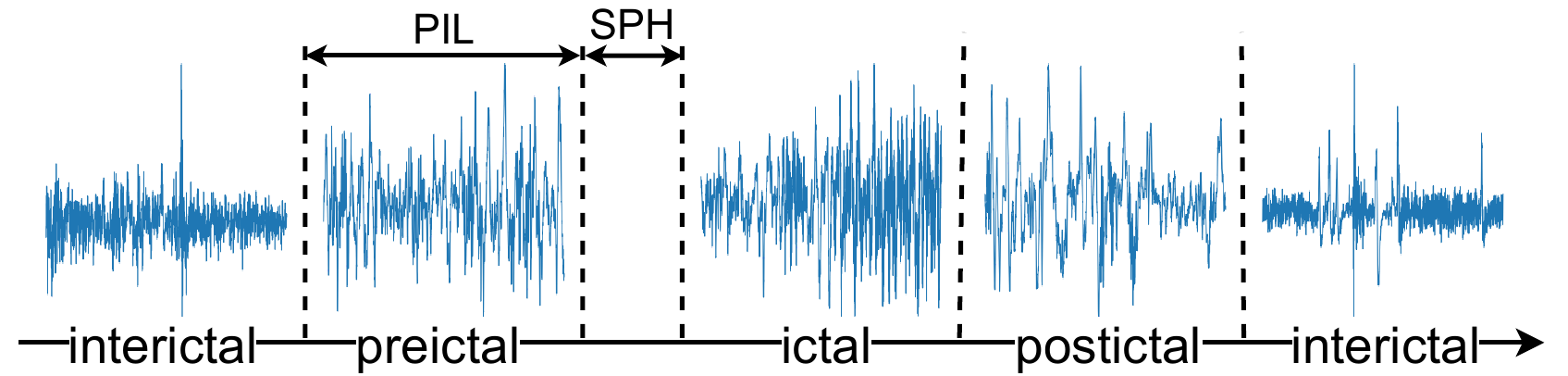}}
\caption{Different states of an epileptic seizure. Seizure prediction horizon (SPH) is a short period between preictal state and seizure onset. The preictal interval length (PIL) is equal to the length of a preictal state. }
\label{fig_seizure_states}
\end{figure}
\section{Results}
\label{sec:results}
\subsection{Dataset and Pre-processing}
\label{sec:data}
\begin{table*}[t]
\centering
\caption{Prediction performance comparison with state-of-the-art methods. N/A stands for results under reconstruction is not applicable since reconstruction is not required to original signal with no compression applied.}
\label{tab:sota_comparison}
\resizebox{\textwidth}{!}{%
\color{black}\begin{tabular}{cccccccc} 
\toprule
Compression ratio                & Metrics          & Lawhern~\cite{Lawhern2018}       & Zhang~\cite{Zhang2021}         & Xu~\cite{Xu2020}            & Truong~\cite{Truong2018}        & This work ($\lambda=0$)         & This work ($\lambda=1$)                           \\ 
\midrule
\multirow{3}{*}{Original signal} & Accuracy (\%)    & 87.2$\pm$1.3  & 89.9$\pm$0.8  & 84.8$\pm$3.7  & 83.4$\pm$1.4  & \textbf{92.5}$\pm$\textbf{1.2}  & N/A                                               \\
                                 & Sensitivity (\%) & 87.9$\pm$2.4  & 93.3$\pm$1.4  & 85.6$\pm$3.4  & 89.7$\pm$2.9  & \textbf{94.2}$\pm$\textbf{1.7}  & N/A                                               \\
                                 & FPR (/h)         & 0.24$\pm$0.03 & 0.14$\pm$0.03 & 0.18$\pm$0.08 & 0.24$\pm$0.05 & \textbf{0.09}$\pm$\textbf{0.02} & N/A                                               \\ 
\midrule
\multirow{3}{*}{$r$ = 1/2}       & Accuracy (\%)    & 85.2$\pm$1.4  & 88.7$\pm$0.9  & 84.3$\pm$3.9  & 83.1$\pm$1.6  & 89.4$\pm$1.4                    & \textbf{90.3}$\pm$\textbf{1.3}                    \\
                                 & Sensitivity (\%) & 88.1$\pm$2.6  & 92.4$\pm$1.3  & 84.7$\pm$3.3  & 88.4$\pm$3.8  & 92.8$\pm$1.6                    & \textbf{93.9}$\pm$\textbf{1.6}                    \\
                                 & FPR (/h)         & 0.25$\pm$0.04 & 0.15$\pm$0.04 & 0.19$\pm$0.10 & 0.25$\pm$0.06 & 0.15$\pm$0.06                   & \textbf{0.13}$\pm$\textbf{0.05}                   \\ 
\midrule
\multirow{3}{*}{$r$ = 1/4}       & Accuracy (\%)    & 85.1$\pm$1.6  & 88.1$\pm$1.2  & 83.7$\pm$3.7  & 81.6$\pm$1.8  & 88.4$\pm$1.6                    & \textbf{89.7}$\pm$\textbf{1.2}                    \\
                                 & Sensitivity (\%) & 87.9$\pm$2.2  & 91.9$\pm$1.4  & 82.6$\pm$3.4  & 86.4$\pm$3.7  & 92.6$\pm$1.7                    & \textbf{\textbf{93.5}}$\pm$\textbf{1.5}           \\
                                 & FPR (/h)         & 0.27$\pm$0.05 & 0.17$\pm$0.04 & 0.24$\pm$0.12 & 0.26$\pm$0.07 & 0.15$\pm$0.08                   & \textbf{0.12}$\pm$\textbf{0.04}                   \\ 
\midrule
\multirow{3}{*}{$r$ = 1/8}       & Accuracy (\%)    & 84.2$\pm$1.7  & 88.3$\pm$1.3  & 82.9$\pm$4.2  & 80.7$\pm$1.6  & 88.6$\pm$1.7                    & \textbf{89.8}$\pm$\textbf{1.2}                    \\
                                 & Sensitivity (\%) & 87.6$\pm$2.5  & 91.7$\pm$1.6  & 81.4$\pm$3.6  & 84.5$\pm$3.8  & 92.8$\pm$1.7                    & \textbf{\textbf{93.4}}$\pm$\textbf{\textbf{1.2}}  \\
                                 & FPR (/h)         & 0.26$\pm$0.11 & 0.18$\pm$0.05 & 0.27$\pm$0.16 & 0.26$\pm$0.09 & 0.19$\pm$0.05                   & \textbf{0.13}$\pm$\textbf{0.03}                   \\ 
\midrule
\multirow{3}{*}{$r$ = 1/16}      & Accuracy (\%)    & 82.5$\pm$1.6  & 87.6$\pm$1.2  & 80.6$\pm$4.6  & 80.2$\pm$1.7  & 88.2$\pm$1.9                    & \textbf{89.7}$\pm$\textbf{1.3}                    \\
                                 & Sensitivity (\%) & 79.8$\pm$2.7  & 90.2$\pm$1.4  & 82.2$\pm$3.5  & 83.3$\pm$4.1  & 90.4$\pm$1.8                    & \textbf{\textbf{93.3}}$\pm$\textbf{1.3}           \\
                                 & FPR (/h)         & 0.29$\pm$0.09 & 0.21$\pm$0.05 & 0.30$\pm$0.14 & 0.27$\pm$0.08 & 0.17$\pm$0.05                   & \textbf{0.13}$\pm$\textbf{0.03}                   \\
\bottomrule
\end{tabular}%
}
\end{table*}
\begin{table}[t]
\caption{Reconstruction performance under different compression ratios.}
\label{tab:reconstruction}
\resizebox{\linewidth}{!}{%
\begin{tabular}{@{}cccccc@{}}
\toprule
Patient ID               & Metrics & $r$ = 1/2 & $r$ = 1/4 & $r$ = 1/8 & $r$ = 1/16 \\ \midrule
\multirow{2}{*}{chb01}   & PSNR   & 40.29    & 42.08    & 40.46    & 35.42     \\
                         & PCC     & 0.97     & 0.97     & 0.96     & 0.91      \\ \midrule
\multirow{2}{*}{chb05}   & PSNR   & 37.40    & 33.21    & 32.06    & 32.68     \\
                         & PCC     & 0.96     & 0.94     & 0.91     & 0.90      \\ \midrule
\multirow{2}{*}{chb06}   & PSNR   & 52.98    & 36.31    & 39.48    & 35.14     \\
                         & PCC     & 0.96     & 0.95     & 0.93     & 0.89      \\ \midrule
\multirow{2}{*}{chb07}   & PSNR   & 36.83    & 32.45    & 31.91    & 31.09     \\
                         & PCC     & 0.92     & 0.81     & 0.82     & 0.79      \\ \midrule
\multirow{2}{*}{chb08}   & PSNR   & 37.95    & 33.46    & 32.89    & 31.64     \\
                         & PCC     & 0.96     & 0.86     & 0.86     & 0.85      \\ \midrule
\multirow{2}{*}{chb09}   & PSNR   & 38.53    & 32.76    & 33.74    & 31.28     \\
                         & PCC     & 0.91     & 0.76     & 0.83     & 0.73      \\ \midrule
\multirow{2}{*}{chb10}   & PSNR   & 44.14    & 40.17    & 39.01    & 37.63     \\
                         & PCC     & 0.98     & 0.96     & 0.94     & 0.93      \\ \midrule
\multirow{2}{*}{chb14}   & PSNR   & 43.55    & 45.92    & 41.87    & 37.38     \\
                         & PCC     & 0.99     & 0.99     & 0.99     & 0.97      \\ \midrule
\multirow{2}{*}{chb18}   & PSNR   & 34.09    & 34.77    & 34.28    & 32.42     \\
                         & PCC     & 0.77     & 0.85     & 0.77     & 0.79      \\ \midrule
\multirow{2}{*}{chb19}   & PSNR   & 39.12    & 39.34    & 39.69    & 35.59     \\
                         & PCC     & 0.93     & 0.93     & 0.91     & 0.87      \\ \midrule
\multirow{2}{*}{chb22}   & PSNR   & 38.83    & 35.50    & 33.94    & 31.62     \\
                         & PCC     & 0.93     & 0.85     & 0.84     & 0.85      \\ \midrule
\multirow{2}{*}{Average} & PSNR   & 40.34    & 36.91    & 36.30    & 33.81     \\
                         & PCC     & 0.93     & 0.90     & 0.89     & 0.86      \\ \bottomrule
\end{tabular}%
}
\end{table}
We evaluate the effectiveness of our proposed framework on the CHB-MIT sEEG database, which is available through open access. The database contains sEEG signals from 23 epileptic patients (17 females, five males, and one person missing gender information). The sEEG signals were collected at a rate of 256 samples per second with 16-bit resolution. Most cases had their sEEG data recorded from 23 channels, and the electrodes were placed according to the International 10-20 system. Each case has a descriptive document, which illustrates relevant information, including case ID, channel information, seizure start time, and end time. The seizure onset time and end time are annotated by clinical experts through visual inspection.

According to the annotation documents, there are channel changes (channels added or removed) in some cases. So we choose the cases that have fixed channels during the acquisition. Furthermore, we are more interested in patients with at least two lead seizures and 1-hour-long preictal time in total. Here, the preictal states are collected only before lead seizures since lead seizures have higher value clinically according to previous work\cite{CHEN202022}. In our study, the lead seizure is defined as a seizure preceded by 4 hours of seizure-free period. We list all patient IDs that qualify our requirements in Table \ref{tab:patient}.

The seizure prediction horizon (SPH) and preictal interval length (PIL) are two critical parameters in determining preictal segments. SPH is a short interval between the end of preictal states and seizure onset. The PIL refers to the duration of preictal states. As is shown in Fig. \ref{fig_seizure_states}, if an alarm occurs at any point within PIL + SPH before seizure onset, it is considered a successful prediction.

The SPH and PIL are still controversial and are usually chosen based on assumptions. SPH offers time for patients to prepare themselves. If the SPH is too large, patients might suffer from anxiety for too long, and the preictal data length might not be sufficient for training; if the SPH is too small, there might not be enough time for the patients to adjust themselves to safe positions.

According to the analysis above, the SPH and PIL are chosen as 5 minutes and 30 minutes respectively in this study, which means the preictal is defined as 5 to 35 minutes ahead of the seizure. In addition, the interictal is defined as 30 minutes after the seizure offset and before the prediction period of subsequent seizure. To reduce the imbalance of preictal and interictal samples, we extract interictal samples from EEG recordings with a sliding window of 20 seconds without overlapping. We apply a 20-second-long sliding window for the preictal samples with 25\% overlapping between two consecutive window sets. We normalize the data by subtracting its mean and dividing its standard deviation. \textcolor{black}{We apply five-fold cross validation on each subject’s data. In specific, we randomly split the data into five equal-sized folds and use four folds as the training set and the remaining one fold as the testing set. The splitting procedure is repeated five times so that each fold is used once as testing set, and we report the mean and standard deviation of all metrics.} We further split 20\% of the training set for validation purpose.

\subsection{Experimental Setup}
\label{sec:exp_setup}
We set the number of filters of the convolution layer of the first up-sampling block in the reconstruction network to be $\#filters_{recon}$ with each following block doubling the number of filters of the previous block. Due to the high variability of EEG signals across different individuals, we train the prediction network in a patient-specific manner. We refer to the number of filters of the stem layer as $\#filters_{stem}$. For each convolution layer of basic convolution block $l$, we set its number of filters to be $2 * l * \#filters_{stem}$. We denote the hidden linear layer size of the prediction network as $size_{fc}$.  For optimization, We adopt the Adam optimizer with learning rate $lr \in \{1e-5, 5e-5, 1e-4, 5e-4, 1e-3, 5e-3, 1e-2\}$ and train each model with a fixed epoch number of 150. Then, for each patient, we sweep over $\#filters_{stem} \in \{4, 8, 16, 32\}$, $size_{fc} \in \{25, 50, 100\}$ and batch size in $\#filters_{stem} \in \{4, 8, 16, 32\}$ to choose the hyper-parameter setup that gives the highest prediction accuracy on the validation set. Our model is implemented using Pytorch framework and trained end-to-end on NVIDIA 2080Ti GPUs for acceleration.

\subsection{Comparison with State-of-the-art}
We evaluate our proposed framework with the following metrics: accuracy, sensitivity, and false prediction rate (FPR) for seizure prediction; Pearson's correlation coefficient (PCC) and peak signal-to-noise ratio (PSNR). Given signal $x_{i}^{c}$ of channel $c$ and time stamp $i$ and its reconstructed version $\hat{x}_{i}^{c}$, PCC and PSNR are defined as follows:
\begin{equation}
PCC = \frac{1}{C} \cdot \sum_{c=0}^{C} \frac{\sum_{i=0}^{N}(x_i^c-\mu^c)(\hat{x}_i^c-\hat{\mu}^c)}{\sqrt{\sum_{i=0}^{N}(x_i^c-\mu^c)^2}\sqrt{\sum_{i=0}^{N}(\hat{x}_i^c-\hat{\mu}^c)^2}},
\label{eq10}
\end{equation}
where $\mu^c$ and $\hat{\mu}^c$ denotes the average value at channel $c$ of original signal and reconstructed signal, respectively.

\begin{equation}
PSNR = 10 \cdot log_{10}(\frac{{\mathrm{max}}(x_{i}^{c})}{MSE}),
\label{eq12}
\end{equation}
where $MSE$ is defined as follows:
\begin{equation}
MSE = \frac{1}{C * N} \cdot \sum_{c=0}^{C}\sum_{i=0}^{N} \norm{x_{i}^{c} - \hat{x}_{i}^{c}}^2,
\label{eq11}
\end{equation}
$C$ is the total number of channels and $N$ is the total length of the signal, $\mathrm{max}$ stands for the maximum value. Higher PSNR and PCC values indicate better reconstruction performance.
We compare the performance of our proposed method with the following deep learning based baselines:
\begin{figure}[t]
\centerline{\includegraphics[width=\columnwidth]{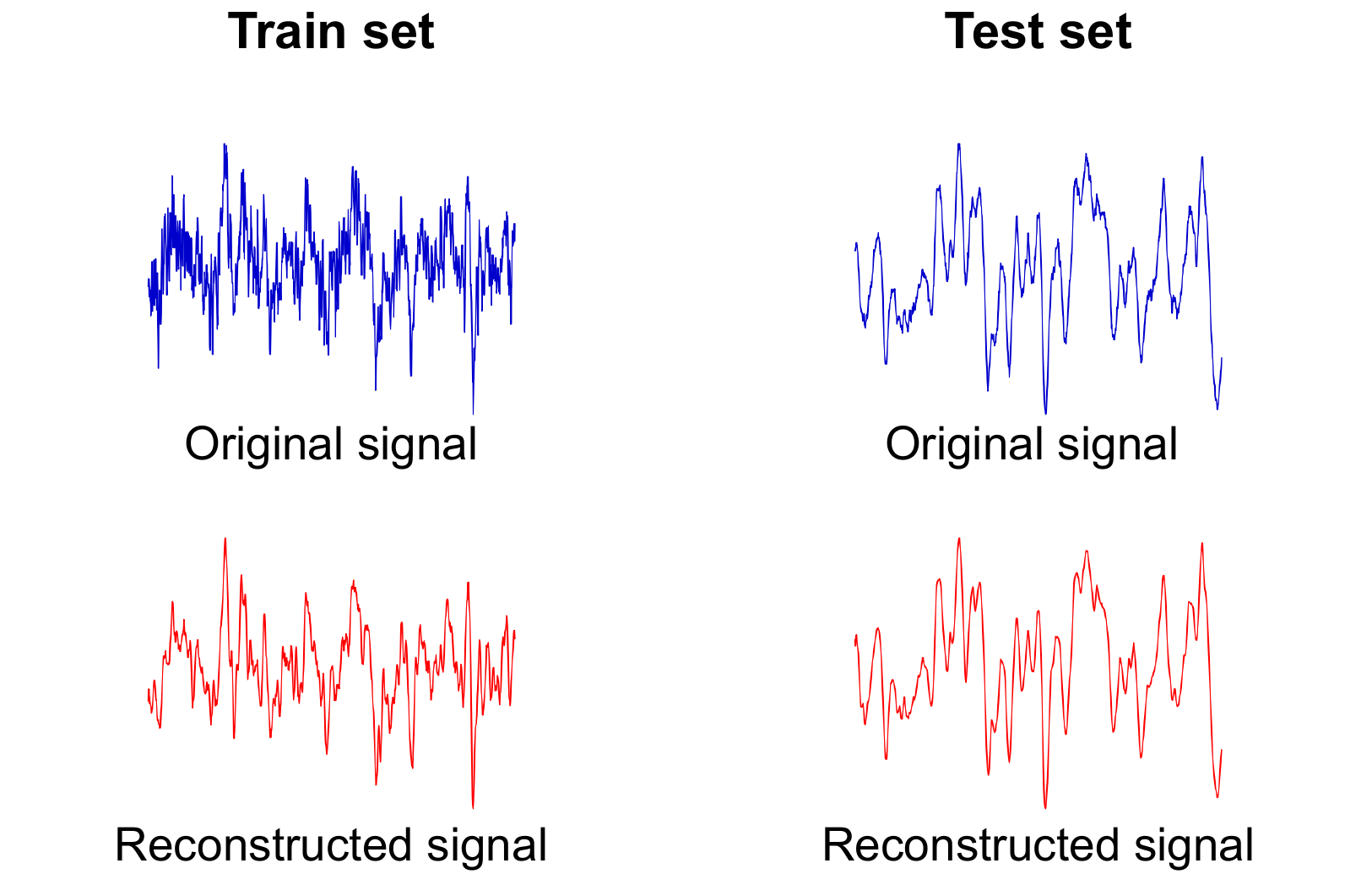}}
\caption{A demonstration of reconstructed signals w.r.t. the original signals of both training and testing set with compression ratio 1/8. }
\label{fig_reconstructed_signal}
\end{figure}
\begin{itemize}
\item Lightweight solution \cite{Zhang2021}: the lightweight solution is based on CNN; it uses synchronization features calculated by Pearson correlation coefficient\cite{00000539-201805000-00050} on all EEG channels as input.
\item End-to-End approach \cite{Xu2020}: the End-to-End patient-specific approach is also based on CNN; it adopts 1-dimensional (1D) kernels in the early-stage convolution and 2D kernels in the late-stage.
\item EEGNet \cite{Lawhern2018}: EEGNet uses compact CNN architecture which contains temporal convolution, depthwise convolution, separable convolution, and pointwise convolution.
\item STFT CNN \cite{Truong2018}: the STFT CNN performs STFT to the EEG signals and performs the classification with a 3-layer CNN.
\end{itemize}

We roughly divide the baselines into two typical types, i.e., methods using original EEG signal as input and methods using statistics of EEG signal as input. Truong \textit{et al.} \cite{Truong2018}, and Zhang \textit{et al.} \cite{Zhang2021} use the result of STFT and PCC of the original signal as the input to the neural network, respectively. We follow the same protocol on the original and compressed signal during comparison.
\begin{figure*}[t]
	\centering
	\includegraphics[scale=0.65]{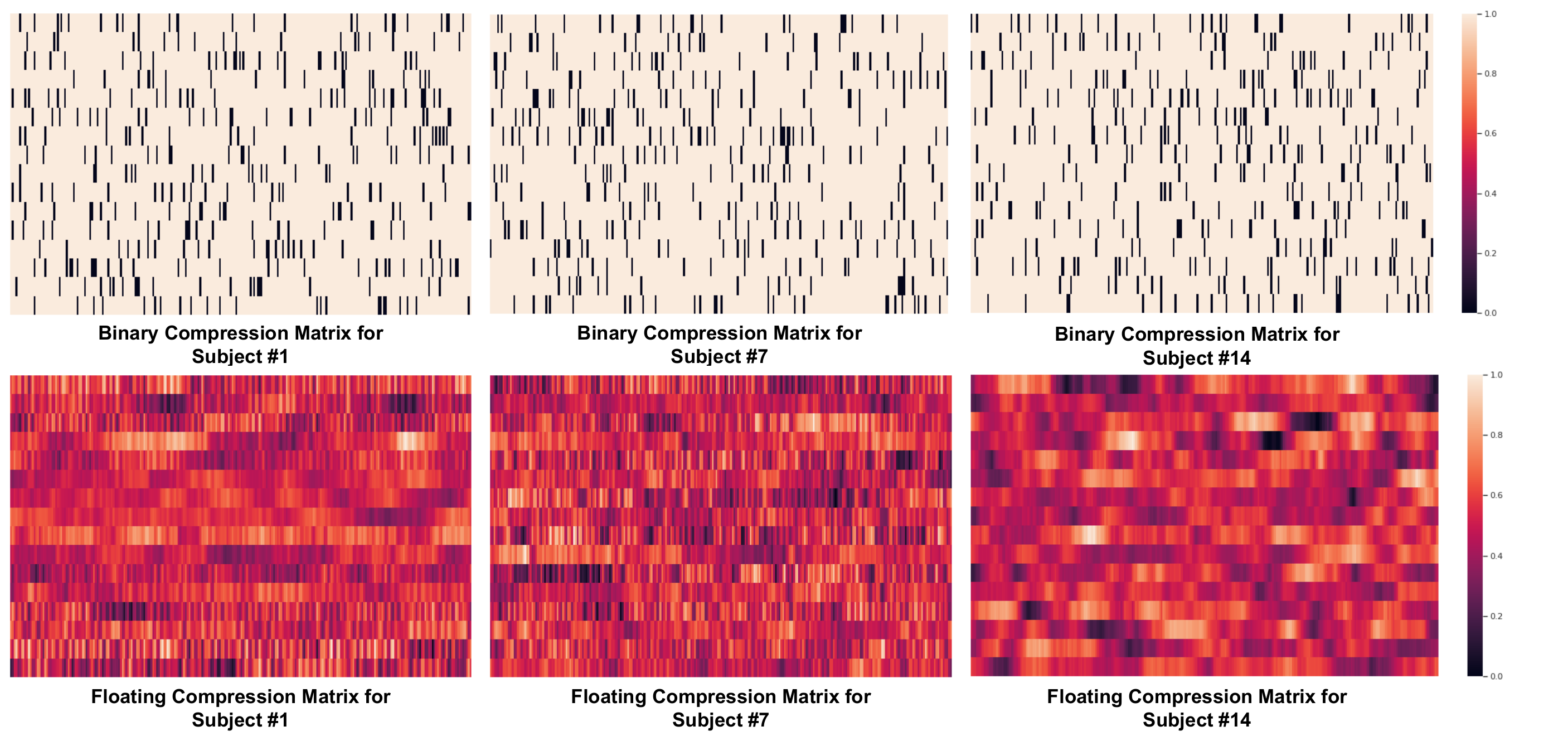}
	\caption{
	\textcolor{black}{Visualization of both floating-point and binary compression matrices for subject one, seven and 14. The first row illustrates the binary matrices while the second show shows the corresponding floating-point compression matrices.}
	}
	\label{fig:matrix}
\end{figure*}
\begin{table*}[t]
\centering
\caption{Prediction performance comparison with state-of-the-art methods using binary compression matrix.}
\label{tab:sota_comparison_b}
\resizebox{\textwidth}{!}{%
\color{black}\begin{tabular}{ccccccc} 
\toprule
Compression ratio                & Metrics          & Lawhern\cite{Lawhern2018}       & Zhang~\cite{Zhang2021}         & Xu~\cite{Xu2020}            & Truong~\cite{Truong2018}        & This work ($\lambda=0$)                                                            \\ 
\midrule
\multirow{3}{*}{Original signal} & Accuracy (\%)    & 87.2$\pm$1.3  & 89.9$\pm$0.8  & 84.8$\pm$1.7  & 83.4$\pm$1.4  & \textbf{92.5}$\pm$\textbf{1.2}                                                     \\
                                 & Sensitivity (\%) & 87.9$\pm$2.4  & 93.3$\pm$1.4  & 85.6$\pm$3.4  & 89.7$\pm$2.9  & \textbf{94.2}$\pm$\textbf{1.7}                                                     \\
                                 & FPR (/h)         & 0.24$\pm$0.03 & 0.14$\pm$0.03 & 0.18$\pm$0.08 & 0.24$\pm$0.05 & \textbf{0.09}$\pm$\textbf{0.02}                                                    \\ 
\midrule
\multirow{3}{*}{$r$ = 1/2}       & Accuracy (\%)    & 84.8$\pm$1.4  & 88.2$\pm$1.0  & 80.2$\pm$1.4  & 83.2$\pm$1.4  & \textbf{90.8}$\pm$\textbf{1.4}                                                     \\
                                 & Sensitivity (\%) & 87.9$\pm$2.6  & 89.9$\pm$2.3  & 86.3$\pm$3.3  & 88.4$\pm$2.8  & \textbf{92.8}$\pm$\textbf{1.6}                                                     \\
                                 & FPR (/h)         & 0.20$\pm$0.04 & 0.21$\pm$0.02 & 0.26$\pm$0.03 & 0.25$\pm$0.07 & \textbf{0.15}$\pm$\textbf{0.06}                                                    \\ 
\midrule
\multirow{3}{*}{$r$ = 1/4}       & Accuracy (\%)    & 80.2$\pm$1.6  & 83.1$\pm$1.2  & 78.7$\pm$1.3  & 83.0$\pm$1.4  & \textbf{90.6}$\pm$\textbf{1.3}                                                     \\
                                 & Sensitivity (\%) & 83.4$\pm$3.2  & 86.4$\pm$2.8  & 86.4$\pm$2.8  & 87.3$\pm$3.8  & \textbf{92.6}$\pm$\textbf{2.3}                                                     \\
                                 & FPR (/h)         & 0.30$\pm$0.05 & 0.30$\pm$0.04 & 0.29$\pm$0.04 & 0.20$\pm$0.05 & \textbf{0.18}$\pm$\textbf{0.04}                                                    \\ 
\midrule
\multirow{3}{*}{$r$ = 1/8}       & Accuracy (\%)    & 77.7$\pm$2.2  & 82.0$\pm$1.4  & 76.7$\pm$1.2  & 80.4$\pm$1.9  & \textbf{89.9}$\pm$\textbf{1.4}                                                     \\
                                 & Sensitivity (\%) & 80.1$\pm$4.7  & 85.5$\pm$2.4  & 79.7$\pm$4.4  & 84.2$\pm$4.9  & \textbf{87.8}$\pm$\textbf{3.1}                                                     \\
                                 & FPR (/h)         & 0.25$\pm$0.06 & 0.25$\pm$0.05 & 0.27$\pm$0.06 & 0.24$\pm$0.07 & \textbf{0.21}$\pm$\textbf{0.05}                                                    \\ 
\midrule
\multirow{3}{*}{$r$ = 1/16}      & Accuracy (\%)    & 76.8$\pm$2.7  & 79.9$\pm$1.5  & 74.6$\pm$1.7  & 78.2$\pm$1.8  & \begin{tabular}[c]{@{}c@{}}\textbf{\textbf{89.4}}$\pm$\textbf{1.5}\\\end{tabular}  \\
                                 & Sensitivity (\%) & 79.3$\pm$4.0  & 83.4$\pm$2.4  & 76.8$\pm$5.0  & 83.8$\pm$4.6  & \textbf{85.8}$\pm$\textbf{2.8}                                                     \\
                                 & FPR (/h)         & 0.27$\pm$0.06 & 0.30$\pm$0.07 & 0.30$\pm$0.05 & 0.29$\pm$0.06 & \textbf{0.25}$\pm$\textbf{0.06}                                                    \\
\bottomrule
\end{tabular}%
}
\end{table*}

To ensure impartial comparison, we closely follow the setup in the original work of the baselines to reproduce their approaches to the best of our effort using Pytorch. All models are trained and evaluated using the same dataset split and pre-processing method as described in Section \ref{sec:data}. We also train each baseline method for 150 epochs and gird search training related hyper-parameters as described in Section \ref{sec:exp_setup}. We adopt a random \textcolor{black}{compression matrix with each element subject to the Gaussian distribution} as the sensing matrix for EEG compression of baseline methods.  To demonstrate the stability of our approach, we report the averaged prediction accuracy, sensitivity, and FPR over all patients listed in Table \ref{tab:patient} of all methods with compression ratios $1/2$, $1/4$, $1/8$ and $1/16$ in Table \ref{tab:sota_comparison}. We also report the result on the original signal for reference. The best performance is marked in bold. As can be seen, our method outperforms all other baseline algorithms in all metrics under compression ratios  $1/2$, $1/4$, and $1/16$, which shows the effectiveness of the proposed framework. We yield slightly lower, but comparable sensitivity than Zhang \textit{et al.} \cite{Zhang2021} with a compression ratio of 1/8. Besides, our model yields the minimum performance variation of 0.6\% in accuracy with compression ratio ranging from 1/2 to 1/16. The two methods \cite{Xu2020, Lawhern2018} using original EEG signal as input suffers more performance drop than methods using statistics\cite{Zhang2021, Truong2018} under compression. It shows that using signal statistics gives more stable performance when using a compressed signal as input. Using a random \textcolor{black}{Gaussian} matrix as a sensing matrix could not capture informative statistics/features for downstream tasks. This explains why the two original signal based methods perform worse. However, using manually extracted statistics as input, on the other hand, discards information embedded in the original signal and thus weakens the deep neural network's feature extraction ability. Our proposed framework kills two birds with one stone by optimizing the sensing matrix together with downstream tasks, which captures informative statistics during compression automatically by the network. Also, Zhang \textit{et al.}'s method outperforms other baseline methods. This performance gain may come from the practice of extracting cross-channel correlation coefficient, which shares a similar idea as adopting the 1D convolution in our proposed prediction network. From Table \ref{tab:sota_comparison}, we also observe that jointly training reconstruction task together with prediction task ($\lambda=1$) yields higher accuracy and sensitivity compared to training prediction task alone ($\lambda=0$). This corresponds to our intuition that the reconstruction task could serve as a regularizer to the prediction task so as to improve the model's generalization ability. To demonstrate the reconstruction performance of our proposed framework, we report PCC and PSNR w.r.t. different compression ratios in Table \ref{tab:reconstruction}. As shown in this table, with a compression ratio of $1/2$, our approach yields an average PSNR of 40.63 and PCC of 0.94. Under compression ratio $1/16$, we observe a reconstruction performance degradation of 6.53 and 0.07 in PSNR and PCC, respectively. This demonstrates the effectiveness of the reconstruction ability of our proposed framework. Besides numerical metrics, we also provide a visualization example of the reconstructed signal of our method in Fig. \ref{fig_reconstructed_signal}. It is observed from the figure that the reconstructed signal of both the training set and test set visually resembles the original signal to a large extent. \textcolor{black}{Next, we consider the case where reconstruction for visualization is negligible, and the main focus is seizure prediction. In this case, we could further reduce the bit precision of the compression matrix to the binary scenario. We show seizure prediction performance without reconstruction under different compression ratios using binary compression matrix in Table \ref{tab:sota_comparison_b}. For baseline methods, we adopt a random compression matrix with each element subject to the Rademacher distribution. As can be seen from Table \ref{tab:sota_comparison_b}, four baseline methods suffer from an accuracy performance degradation of 8.9\% on average with a signal compression ratio of 1/16. Our proposed method only shows a degradation of 3.1\% in accuracy with a signal compression ratio of 1/16. Compared to using a random Gaussian matrix as compression matrix, baseline methods demonstrate a drastic performance degradation as the compression ratio ranges from 1/2 to 1/16. On the contrary, our proposed method shows similar seizure prediction performance when using floating-point and binary compression matrices. This further proves that learning compression matrices along with the downstream task could embed informative features into the compressed signal and thus yield an ideal trade-off between seizure prediction performance and signal transmission power consumption. Finally, we demonstrate a visualization of both the floating-point and binary compression matrices learned with a compression ratio of 1/16 for different patients in Figure \ref{fig:matrix}. The first row shows the binary compression matrices for subjects one, seven, and 14, while the second row shows the corresponding floating-point compression matrices. It is observed from Figure \ref{fig:matrix} that the compression matrices vary significantly across patients no matter the bit precision. This proves that our proposed framework captures subject-specific informative features during the compression process, which further improves seizure prediction performances.}

\section{Conclusion}
\label{sec:conclusion}
We proposed in this paper a novel learning framework aiming at designing an efficient and reliable seizure prediction system with compressed EEG signals. The proposed approach jointly optimizes compression, reconstruction, and prediction tasks in an online fashion. Specifically, we implemented a CNN network with residual connection for prediction and reconstruction network with adaptive architecture. The learned compression matrix thus captures informative features for reconstruction and prediction during the compression process. Extensive experiments over a benchmark dataset show that the proposed approach outperforms not only state-of-the-art methods by a large margin but also indicates low degradation in prediction accuracy under high compression. After training, the compression matrix could be deployed in wearable devices for stable and reliable seizure prediction. Our proposed framework could be easily extended to other implanted applications such as ECoG-based measurements. Future work includes refining our algorithms to patient-independent prediction.

{\small
\bibliographystyle{ieeetr}  
\bibliography{ref}
}
\end{document}